# Efficient Antenna Optimization Using a Hybrid of Evolutionary Programing and Particle Swarm Optimization

Ahmad Hoorfar and Shamsha Lakhani

Antenna Reserach Laboratory, Department of Electrical and Computer Engineering,
Villanova University, Villanova, PA, 19085
email: ahoorfar@villanova.edu

*ABSTRACT*

*In this paper, we present a hybrid of Evolutionary Programming (EP) and Particle Swarm Optimization (PSO) algorithms for numerically efficient global optimization of antenna arrays and metasurfaces. The hybrid EP-PSO algorihm uses an evolutionary optimization approach that incorporates swarm directions in the standard self-adaptive EP algorithm. As examples, we have applied this hybrid technique to two antenna problems: the side-lobe-level reduction of a non-uniform spaced (aperiodic) linear array and the beam shaping of a printed antenna loaded with a partially reflective metasurface. Detailed comparisons between the proposed hybrid EP-PSO technique and EP-only and PSO-only techniques are given, demonstrating the efficiency of this hybrid technique in the complex antenna design problems.*

## 1. INTRODUCTION

Applications of the evolutionary and multi-agent stochastic optimization techniques such as Evolutionary Programming (EP) [1, 2], Genetic Algorithms (GAs) [3, 4], Particle Swarm Optimization [5], and Covariance Matrix Evolution Strategy (CMA-ES) [6] in electromagnetic and antenna design have demonstrated that these probabilistic methods can yield robust globally optimized solutions to high-frequency design problems that otherwise are not amenable to traditional gradient-based local-search optimization methods [7-12]. A comparative study of various variations of EP, GAs, PSO to various antenna problems was reported in [13] where it was shown that EP has the best performance among these algorithms, whereas PSO provides a few very high-quality solutions, which are comparable to or better than those of EP, majority of its solutions over a large number of trials are rather poor as compared to other algorithms. The questions naturally arise as to whether a judicious hybridization of these algorithms may result in a new algorithm with a faster convergence in typical antenna optimization problems.

In this work we investigate the efficiency of an evolutionary optimization approach that incorporates swarm directions in standard EP algorithm. Our hybrid EP-PSO approach is inspired by the work reported in [14] for test-function optimization. Unlike [14] however, we have also investigated the use of various mutation operators in the hybrid EP-PSO technique, as well as the optimum selection of boundary conditions in PSO for a given antenna problem when the proposed hybrid scheme is applied. As examples, we have applied the technique to two antenna problems: the side-lobe-level reduction of a non-uniform spaced (aperiodic) linear array and the beam shaping of a printed antenna loaded with a partially reflective metasurface. Detailed comparisons between the hybrid EP-PSO technique and EP-only and PSO-only techniques are given, demonstrating the efficiency of the hybrid technique in the investigated antenna design problems. Some results of this work were previously presented in a conference [15].

## 2. FORMULATION OF HYBRID EP-PSO TECHNIQUE

Considering a solution vector of size n, $\bar{x}_i = [x_i(1), x_i(2), \ldots, x_i(n)]$, in standard EP algorithm with self-adaptive Gaussian mutation, offspring's and their corresponding strategy parameter, $(\bar{x}_i', \bar{\eta}_i')$ are generated from a parents population, $(\bar{x}_i, \bar{\eta}_i)$, $\forall i \in \{1, \ldots, \mu\}$, according to [16]:

$$x_i'(j) = x_i(j) + \eta_i'(j) N_j(0,1) \tag{1}$$

$$\eta_i'(j) = \eta_i(j) e^{[\tau' N(0,1) + \tau N_j(0,1)]} \tag{2}$$

with

$$\tau = \left(\sqrt{2\sqrt{n}}\right)^{-1}, \tau' = \left(\sqrt{2n}\right)^{-1} \tag{3}$$

for j = 0,1,2,....n, where *x(j) and η(j)* and are the jth components of the solution vector and the variance vector, respectively. N(0,1) denotes a one-dimensional random variable with a Gaussian distribution of mean zero and standard deviation one. Nj(0,1) indicates that the random variable is generated anew for each value of j. In the hybrid EP-PSO approach we simply modify (1) and incorporate swarm directions in it according to:

$$v_i(j) = c_1 R(0,1) \left[ x_{igroup}(j) - x_i(j) \right] + c_2 R(0,1) \left[ x_{ipast}(j) - x_i(j) \right] \tag{4}$$

$$x_i(j) = x_i(j) + v_i(j) \tag{5}$$

$$x_i'(j) = x_i(j) + \eta_i'(j) N_j(0,1) \tag{6}$$

Where $x_{igroup}$ and $x_{ipast}$ are the best member position (here the one with minimum fitness value in the population) and the member's previous best position, respectively, R(0,1) is a random number with a uniform distribution in the range [0,1], and c1 and c2 are the cognitive and social rate coefficients in the standard PSO algorithm. In this approach, the particles position is first updated by (4)-(5), and then the offspring population are generated in the new position by the standard EP mutation step in (6). We note that one may also replace Gaussian mutation in (6) with Cauchy or an adaptive hybrid of Cauchy and Gaussian mutation operator, which has been shown to be more efficient in certain electromagnetic optimization problems [13]. This is done by replacing equation (6) with [25]:

$$\bar{x}_i'(j) = \bar{x}_i(j) + \bar{\eta}_i'(j) C_j(0,1) \tag{7}$$

where C (0,1) denotes a random variable with Cauchy distribution centered at the origin and with the scale parameter equal to one.

For other steps in the optimization process we follow the initialization, selection and tournament process outlined in [16]. The flow-chart of our implementation of the proposed hybrid EP-PSO is shown below in Figure 1.

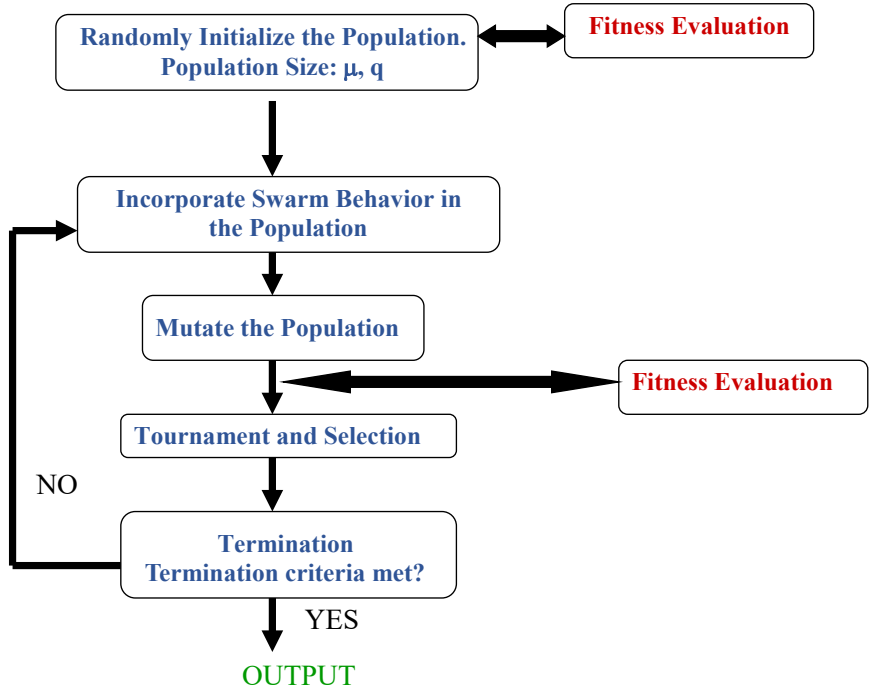

Figure 1: Flowchart of the Hybrid EP-PSO algorithm

## 3. OPTIMIZATION OF NON-UNIFORMLY SPACED ARRAYS

As the first application of the proposed technique, the hybrid EP-PSO algorithm was applied to the optimization of the maximum side-lobe level (SLL) of a 20 element non-uniformly spaced array of isotropic elements with the total aperture size of 8 λ, shown in Figure 2.

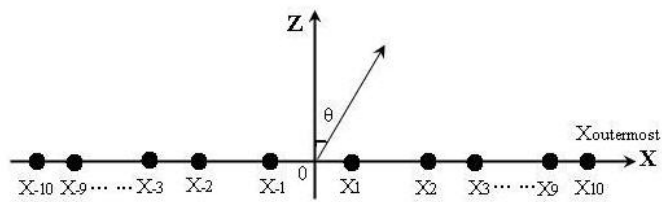

Figure 2: A non-uniformly spaced Array

Assuming a symmetric array, the location of the outermost element, $x_{10}$, was specified to $x_{10} = 4\lambda$. The locations of other elements were optimized, resulting in a 9 parameter optimization problem. The solution vector was defined as:

$\mathbf{X} = [x_1, x_2, x_3, x_4, \ldots x_9]$; $x_i \in [0, 4\lambda]$, $|x_i - x_j| > 0.3\lambda$; and $\min\{x_i\} > 0.15\lambda$, $i=1,2,3,4\ldots 9$, $i \neq j$.

The fitness value (objective function) for this problem was famulated to be equal to the maximum sidelobe level, *Fitness Value* = $SLL_{max}$ where $SLL_{max}$ is the value of the largest side-lobe level of the array.

### A. Comparison of Various Algorithms in Non-Uniformly Spaced Array Design

First, to assess the performance of various standard forms of EP, GA, micro-GA [17], and PSO, those algorithms were applied to this array problem. The strategy parameters for each algorithm were selected based on the optimal parameters for this problem given in [13]. For PSO, the invisible boundary was used as it, followed very closely by the absorbing boundary, had been shown to provide faster convergence in the antenna array optimizations [13]. The inertia coefficient in PSO was set to 0.7, and the cognitive and the social rate coefficients were set to 1.0 and 2.0, respectively. For EP, the EP algorithm with Gaussian mutation operator, EP-GMO, was used as it provides better convergence than the EP algorithm with Cauchy mutation operator, EP-CMO, for this problem [13]. For both GA and micro-GA a uniform crossover was used, and the crossover and mutation probabilities for GA were fine-tuned to Pc=0.7 and Pm=0.02, respectively, crossover probability for micro-GA was set to Pc=0.9 [13].

A population size of 50 individuals or particles in each generation and 200 generations per trial were used in all algorithms. For the micro-GA, 6 members (5 parents and 1 elite member) in each generation were used. If the gene difference between the best (elite) member and all other member was less than 20%, all members for the next generation, except for the best member, were re-generated. Figures 3 and 4 compare the convergence rates of the four algorithms over 100 trials in optimization of an aperiodic array of isotropic elements. In addition, included for comparison is the optimization using PSO with a member size of 100. The overall best solution of PSO has the fastest convergence among all the algorithms for the first 3,000 fitness evaluations, and its performance is comparable to that of EP as the number of fitness evaluations increases. As shown in Figure 4, however, when averaged over the 100 trials, EP easily outperforms both GAs and PSO, with a rather poor performance by the latter.

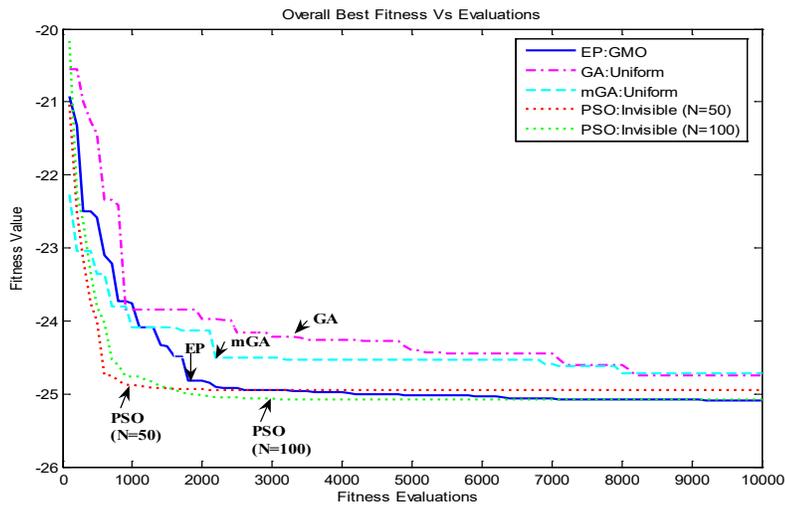

Figure 3: Best-solution fitness trajectories of EP, GA, micro-GA and PSO convergence in optimization of non-uniform array over 100 trials

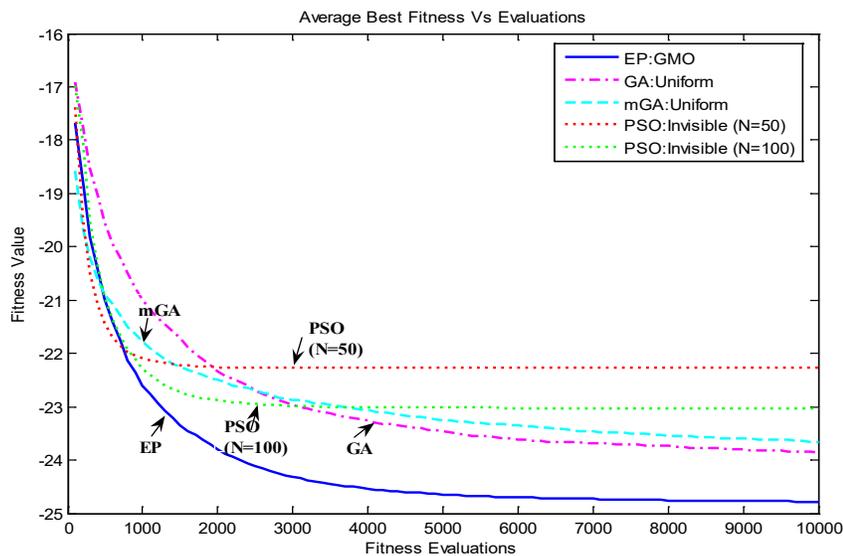

Figure 4: Average-best fitness trajectories of EP, GA, micro-GA and PSO convergence in optimization of non-uniform array, averaged over 100 trials.

*B. Application of Hybrid EP- PSO to Non-Uniformly Spaced Array*

In the above subsection 3-A, we demonstrated that EP convincingly provides the fastest overall convergence among the four algorithms utilized for the problem of non-uniformly spaced array. Now to assess the efficiency of the hybrid EP-PSO (HEPPSO) algorithm, we compare its performance with EP with Gaussian mutation and PSO with absorbing wall boundary. These versions of EP and PSO were chosen for comparison, since, as we previously discussed, they had been demonstrated to perform better than the other versions of EP and PSO for optimization of non–uniformly spaced array [13, 18].

For comparison, a population size of 50 was used for 200 generations and 20 trials. The average fitness trajectories of the best solution member along with the pattern of the final optimized solution are given in Figures 5 and 6. The coefficients c1 and c2 in equation (3) were set equal to 2, the number of opponents in the EP tournament process were set to $q = 15$ for both the EP and Hybrid EP-PSO algorithms, It can be seen from the average best fitness trajectory that the Hybrid EP-PSO outperformed the EP only and PSO only algorithms. The pattern corresponding to the best optimized design has a maximum side-lobe level of less than -25 dB. The optimized spacing distances, in free-space wavelength, between the elements were: $X_{01} = 0.16347$ $X_{12} = 0.30039$, $X_{23} = 0.30095$, $X_{34} = 0.35352$, $X_{45} = 0.31826$, $X_{56} = 0.37901$, $X_{67} = 0.41163$, $X_{78} = 0.45817$, $X_{89} = 0.59966$.

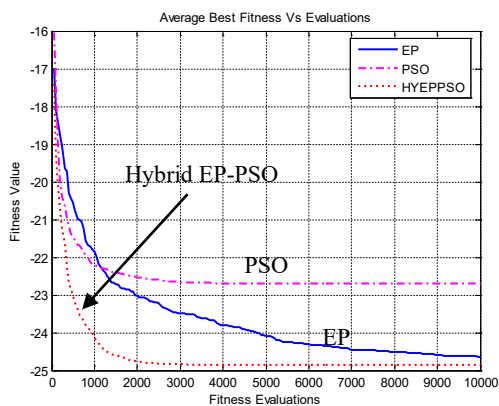
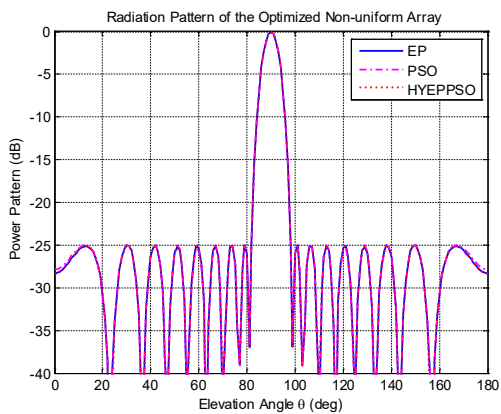

Figure 5: Average best fitness trajectory over 20 trials

Figure 6: Best optimized pattern for a 20-element Non–uniformly spaced array

## 3. OPTIMIZATION OF METASURFACES FOR ANTENNA BEAM-SHAPING

A class of metasurfaces known as partially reflective surfaces (PRS) have been used in antenna applications for beamforming [19] as well as more for high gain wireless applications [20]. A simple implementation of the PRS for the latter application includes a periodic array of narrow printed strips with uniform spacing placed in front of a printed dipole or a microstrip patch antenna at a resonant distance of about 0.5 wavelength. Such a uniformly spaced array of PRS elements, however, results in a relatively high sidelobe level. It also lacks sufficient degrees of freedom, when fed by a printed antenna, to optimize the resulting radiation characteristics for a prescribed shaped beam pattern. To overcome this, non-uniformly spaced metasurfaces have been proposed when the spacing between the PRS elements can be used as additional optimization parameters [21, 22].

We have applied Evolutionary Programming, Particle Swarm Optimization and Hybrid EP-PSO to optimize non-uniformly partially reflective surfaces, shown in Figure 7. These algorithms were linked to a Method of Moment code for multilayered printed structures, developed in-house based on a mixed-potential integral equation formulation [23, 24], in order to numerically model the antenna structure. This numerical engine accounts for all the coupling effects in design of optimization of the non-uniform PRS-based antenna structure in Figure 7. In this design, a partially reflective surface (PRS) as shown in Figure 7 is a group of narrow conducting strip elements placed in front of a feeder antenna at a distance 'd'.

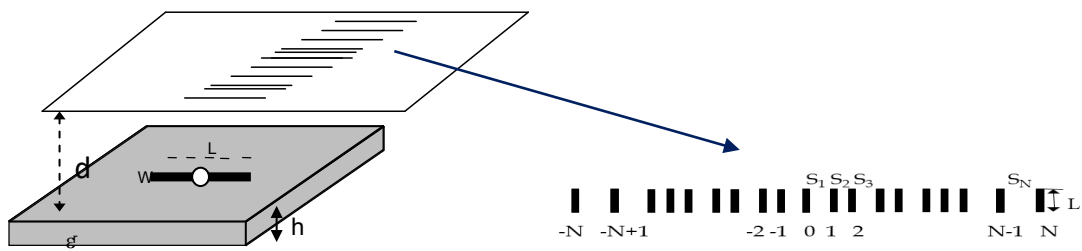

Figure 7: A metasurface made of non-uniformly spaced PRS elements array fed by microstrip dipole

The objective function formulated as:

$$F(\bar{x}) = \sum_{0}^{M_\theta} \left\| E_{nor}(\theta, \varphi; \bar{x}) \right|_{dB} + \sum_{m=0}^{M_\theta} r(m) \left[ \delta - \left\| E_{nor}(\theta_m, \varphi; \bar{x}) \right|_{dB} \right] + \alpha (\left| SLL \right|_{dB} - \left| SLL_{max} \right|_{dB}) \quad (8)$$

where

$$r(m) = \begin{cases} q_1, & \left\| E_{nor}(\theta_m, \phi, \bar{x}) \right|_{dB} > \delta \\ 0, & otherwise \end{cases} \quad ; \quad \alpha = \begin{cases} q_2, & \left| SLL \right|_{dB} > \left| SLL_{max} \right|_{dB} \\ 0, & \left| SLL \right|_{dB} \leq \left| SLL_{max} \right|_{dB} \end{cases} \quad (9)$$

with the solution vector given by:

$$\bar{X} = \left[ S_1, S_2, ..., S_N; L_1, L_2, ..., L_N; W_p, L_p; d \right]^T \quad (10)$$

In (8) and (9), $|E_{nor}(\theta, \phi)|$ is the normalized pattern and $M_\theta$ is the number of elevation angles, $\theta$, sampled in the interval [0, $\theta_{max}$], in which we require a near flat-top pattern, with a maximum dB deviation of $\delta$, in $\phi$-plane, $\phi = 0$ or 90 degrees' plane. The third term in (8) penalizes all the solutions that violate the constraint on the maximum sidelobe-level in both planes.

We have optimized a 21-element partially reflective surface for the synthesis of 120 degrees of sectoral (flat top) pattern at 26.35GHz. A centre-fed printed dipole of length 5.69mm on a 0.254mm air substrate was used for excitation of the PRS elements. Strip width of 0.28mm were used as PRS elements. The spacing, $S_i$, i=1,2,…, 10, between the PRS elements, length of the elements and distance from the dipole were optimized to get a flat top pattern. A symmetric PRS structure was assumed.

For all the three algorithms a population size of 20 was used. For EP and Hybrid EP-PSO 15 opponents were used. The coefficients c1 and c2 were set equal to 2 for PSO and Hybrid EP-PSO.

To use the optimal choice of boundary wall for PSO, first a study was performed to assess the performance of PSO for absorbing, invisible and reflecting walls. The results for the mean best trajectories, averaged over 20 trials, for the three boundary cases are shown in Figure 8. As can be seen, the PSO with absorbing wall performs the best for this metasurface example. Hence the absorbing wall condition was also used in the hybrid EP-PSO algorithm for this problem.

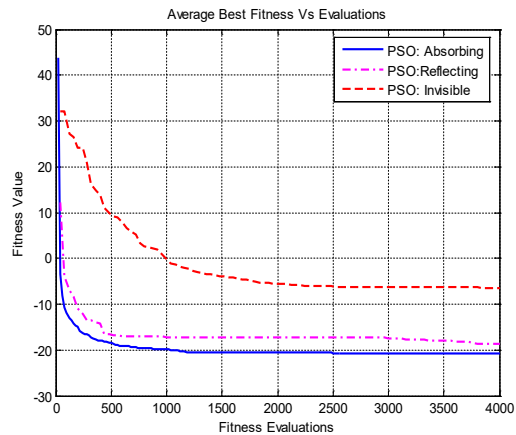

Figure 8: Trajectory of the best solution member using PSO averaged over 20 trials for different boundary walls

Figure 9 shows the mean best fitness averaged over 20 trials for EP, PSO, and hybrid EP-PSO. The final optimized pattern, optimized for a 120 degree flattop pattern with a constraint of $\delta = 0$dB, in $\phi = 0°$ plane, is shown in Figure 10. As can be seen, all three algorithms eventually result in the optimized pattern shown in Figure 10, but the hybrid EP-PSO converges on the average after about 400 fitness evaluations as compared to about 800 evaluations for the other two algorithms. The final optimized dimensions of the structure obtained using the hybrid EP-PSO algorithm are summarized in Table 1.

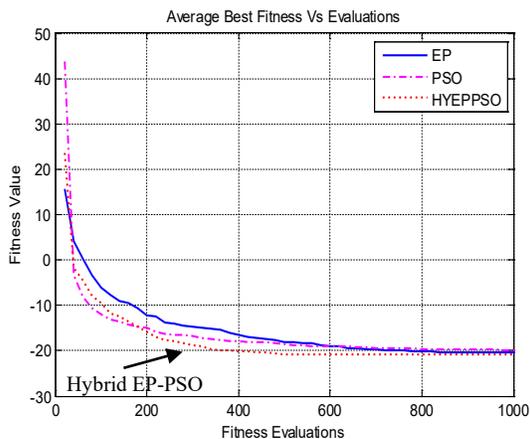
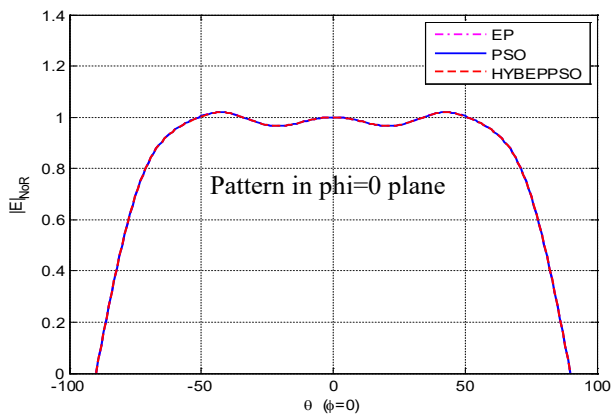

Figure 9: Average best fitness trajectory over 20 trials in optimization of PRS structure

Figure 10: Best optimized pattern for a 21-element Non–uniformly spaced PRS structure

Table 1: Optimized dimensions: $S_1$ through $S_{10}$ are the spacings between PRS elements, L is the length of PRS elements and d is the distance of the metasurface from the microstrip dipole.

| Parameters (mm) | S1 | S2 | S3 | S4 | S5 | S6 | S7 | S8 | S9 | S10 | L | d |
|---|---|---|---|---|---|---|---|---|---|---|---|---|
| Optimized dimensions | 0.5475 | 0.4638 | .4796 | .5933 | .6071 | .70 | .70 | .70 | .6938 | .70 | 6.5455 | 8 |

## 4. CONCLUSIONS

In this paper, we outlined the formulation of a hybrid of EP and PSO algorithms that incorporates swarm directions in standard self-adaptive EP algorithm and applied it to design optimization of two antenna problems: side-lobe-level minimization of a non-uniformly spaced linear array of isotropic radiators and the pattern synthesis of a partially reflective metasurface fed by a microstrip dipole to obtain a near flat-top pattern. For each problem, a study was first performed to determine the optimum strategy parameters for best statistical performance of the utilized EP and PSO algorithms. It was shown that in both cases the hybrid EP-PSO algorithm outperform the EP only and PSO only algorithms in these two antenna problems that have significantly different fitness function landscapes.